\documentclass{article}
\usepackage{ijcai21}
\usepackage{times}

\usepackage{soul}
\usepackage{url}
\usepackage[hidelinks]{hyperref}
\usepackage[utf8]{inputenc}
\usepackage[small]{caption}
\usepackage{graphicx}
\usepackage{amsmath}
\usepackage{booktabs}
\usepackage{subfig}
\usepackage{multirow}
\usepackage{amsmath}
\usepackage{amsfonts} 
\usepackage{nicefrac} 
\usepackage{microtype} 
\usepackage{amsthm}
\usepackage{algorithm}
\usepackage{algorithmic}
\usepackage{xcolor}
\def \R {\mathbb{R}}

\def \R {\mathbb{R}}

\def \W {\mathcal{W}}

\DeclareMathOperator*{\argmin}{arg\,min}

\def\hlinewd#1{%
\noalign{\ifnum0=`}\fi\hrule \@height #1 %
\futurelet\reserved@a\@xhline} 

\title{Deep Descriptive Clustering}

\author{
Hongjing Zhang\and
Ian Davidson\\
\affiliations
University of California, Davis\\
\emails
hjzzhang@ucdavis.edu,
davidson@cs.ucdavis.edu
}

\begin{document}

\maketitle

\begin{abstract}
Recent work on explainable clustering allows describing clusters when the features are interpretable. However, much modern machine learning focuses on complex data such as images, text, and graphs where deep learning is used but the raw features of data are not interpretable. This paper explores a novel setting for performing clustering on complex data while simultaneously generating explanations using interpretable tags. We propose deep descriptive clustering that performs sub-symbolic representation learning on complex data while generating explanations based on symbolic data. We form good clusters by maximizing the mutual information between empirical distribution on the inputs and the induced clustering labels for clustering objectives. We generate explanations by solving an integer linear programming that generates concise and orthogonal descriptions for each cluster. Finally, we allow the explanation to inform better clustering by proposing a novel pairwise loss with self-generated constraints to maximize the clustering and explanation module's consistency. Experimental results on public data demonstrate that our model outperforms competitive baselines in clustering performance while offering high-quality cluster-level explanations.
\end{abstract}

\section{Introduction}
As machine learning is applied to more complex data and situations, the need to understand a model's decisions becomes more paramount. The area of explainable AI (XAI) \cite{adadi2018peeking} is motivated to enhance the interpretability of complex machine learning models, especially deep learning.
Arguably XAI is more needed and more challenging in unsupervised learning such as clustering as the explanations are usually at the model level rather than the instance level. For example, supervised learning explanations mainly focus on why an instance is classified to a specific class \cite{ribeiro2016should}; however, with clustering we need to explain the semantics of each discovered cluster. 

\begin{figure}[t]
\centering
 \includegraphics[width=0.45\textwidth]{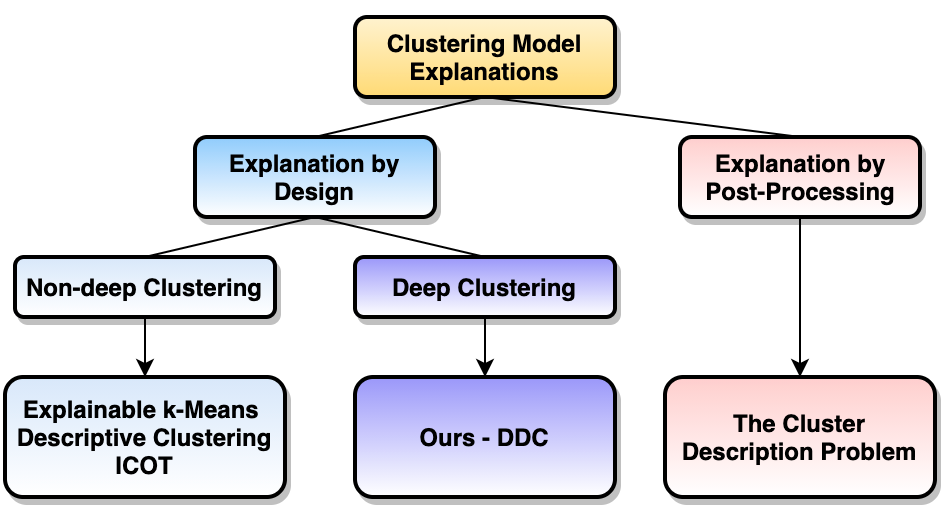}
\caption{Taxonomy of works on clustering explanations.}
\vskip -0.1in
 \label{fig:problem_setting}

\end{figure}

Existing work on explainable clustering (see Figure \ref{fig:problem_setting}) typically takes one of two directions: i) explanation by design algorithms \cite{bertsimas2020interpretable,moshkovitz2020explainable} that use interpretable features to create both a clustering and an explanation (left branch of Figure \ref{fig:problem_setting}). ii) \textcolor{black}{explanation by post-processing \cite{davidson2018cluster,Sambaturu_Gupta_Davidson_Ravi_Vullikanti_Warren_2020} which take an existing clustering and generate an explanation using another additional set of features (tags)  not given to the clustering algorithm (right branch of Figure \ref{fig:problem_setting})}. Each direction has its limitations: the first type of work is not suitable for complex data such as images and graphs as the underlying features are uninterpretable to a human. Post-processing methods are algorithm-agnostic, but they did not fully use the information from additional features to guide the clustering process hence may generate poor explanations as the clustering may be difficult to explain. Instead, our proposed method will learn a good clustering that is also interpretable while post-processing approaches only find the best explanation possible for a given clustering.

Explaining deep clustering results using the underlying complex features is a challenging but alternative direction. \textcolor{black}{Instead, we explore the situation that a partial set of instance-level semantic tags are known from which we generate a cluster-level description along with complex features to cluster on.} This setting is not uncommon and was previously studied \cite{dao2018descriptive,davidson2018cluster}. Example settings include Twitter graphs where the user-user graph is clustered and explained using hashtags and personal images where the tags are descriptors of events in the image.

To address the challenges of simultaneously clustering and explaining efficiently with incomplete semantic features, we propose a novel deep descriptive clustering framework (DDC) that incorporates both deep clustering and cluster-level explanation. The whole framework is trained iteratively to maximize the consistency between the clustering and its explanation. To be specific, we formulate the cluster-level description problem as an Integer Linear Programming (ILP) which is solved for concise and orthogonal descriptions for each cluster.  Inspired by the success of the discriminative clustering model \cite{krause2010discriminative} which has fewer assumptions of the data, our main clustering objective maximizes the mutual information between empirical distribution on the inputs and the induced clustering labels.
Finally, we propose a pairwise loss with self-generated constraints to penalize the inconsistency between the clustering feature space and discovered descriptive tag space to improve the clustering quality. The major contributions of this paper are listed as follows:  
\begin{itemize}
    \item We propose a novel framework to learn clustering and cluster-level explanations simultaneously. The proposed architecture supports learning from the sub-symbolic level (which clustering is performed on) and symbolic level (which explanations are created from). 

\item We formulate the class-level explanation problem as an ILP to solve concise and orthogonal explanations. A pairwise loss function is proposed with self-generated constraints to bridge the clustering and explanation.
   \item Empirical results on public data demonstrate that our model consistently achieves better clustering results and high-quality explanations compared to recent baselines. 
   \item We explore the novel direction of graphical ontology explanations for clustering when the number of clusters is large and a lengthy explanation list is problematic.
\end{itemize}

The rest of this paper is organized as follows. In section \ref{sec:related_work}, we overview related works. We then introduce our learning objectives and optimization algorithms in section \ref{sec:approach}. Experimental results and analysis are reported in section \ref{sec:experiments}. Finally, section \ref{sec:conclusion} concludes this paper with future research directions.

\section{Related Work}
\label{sec:related_work}

\paragraph{Explainable clustering models.} Many previous works \cite{liu2005clustering,fraiman2013interpretable,ghattas2017clustering,bertsimas2020interpretable} have explored the explainable-by-design algorithms which consider the simultaneous construction of decision trees and cluster discovery for explainable clustering. Typical work such as \cite{moshkovitz2020explainable} considered traditional clustering algorithms like k-medians/means. However, one major limitation of these methods is that they require the features used for clustering to be interpretable which may not be the case for complex data such as graphs and images. Another line of research \cite{davidson2018cluster,Sambaturu_Gupta_Davidson_Ravi_Vullikanti_Warren_2020} assumes one set of semantic tags for each instance are available to do post-processing explanation. \cite{davidson2018cluster} proposed a model-agnostic explanation method that explains any given clustering with tags but does not change the clustering. Perhaps the closest work to our own is \cite{dao2018descriptive} but is limited to simple diameter-based clustering objectives and scales to just a few thousand instances whilst making strong assumptions such as having well-annotated tags for every instance. Our work differs from the above: we learn a \emph{deep clustering} model and cluster explanation \emph{simultaneously} with a \emph{partial} set of semantic tags and scales for \emph{large} data sets.

\paragraph{Multi-view clustering.} As our approach uses semantic tags for explanation this can be seen as another view of the data; hence we overview the recent works on multi-view clustering and discuss how our proposed work differentiates from it. The goal of multi-view clustering \cite{bickel2004multi,xu2013survey,shao2015multiple,tao2017ensemble,wang2018multiview,hu2018doubly} is getting better clustering results via exploiting complementary and consensus information across multiple views rather than simply concatenating different views. Our descriptive clustering setting is different from multi-view clustering: Firstly, instead of just one goal which maximizes clustering performance, our work has another explanation objective to find meaningful descriptions of clusters. Secondly, most multi-view clustering is for similar views (i.e., all images) whereas our views are more diverse (e.g., continuous image features with categorical semantic tags) than general multi-view clustering settings.

\paragraph{Constrained clustering.} Unlike most multi-view clustering algorithms which leverages the knowledge from different views to maximize the clustering performance, constrained clustering assumes the users have access to partial pre-existing knowledge about the desired partition of the data. The constraints are usually expressed via pairwise constraints \cite{wagstaff2001constrained,bilenko2004integrating,basu2008constrained} such as \emph{together} and \emph{apart} which indicates whether two instances belong to the same cluster or different clusters. Recent works \cite{fogel2019clustering,zhang2019framework} have also extended constrained clustering to deep learning models. Our work shares one common attribute with these works in using a constrained optimization objective for better clustering. However, in this work our constraints are dynamically self-generated in that they cannot be known a priori as generating those constraints require both the feature representation and the clustering explanations.

\begin{figure*}[th]
\centering
 \includegraphics[width=0.86\textwidth]{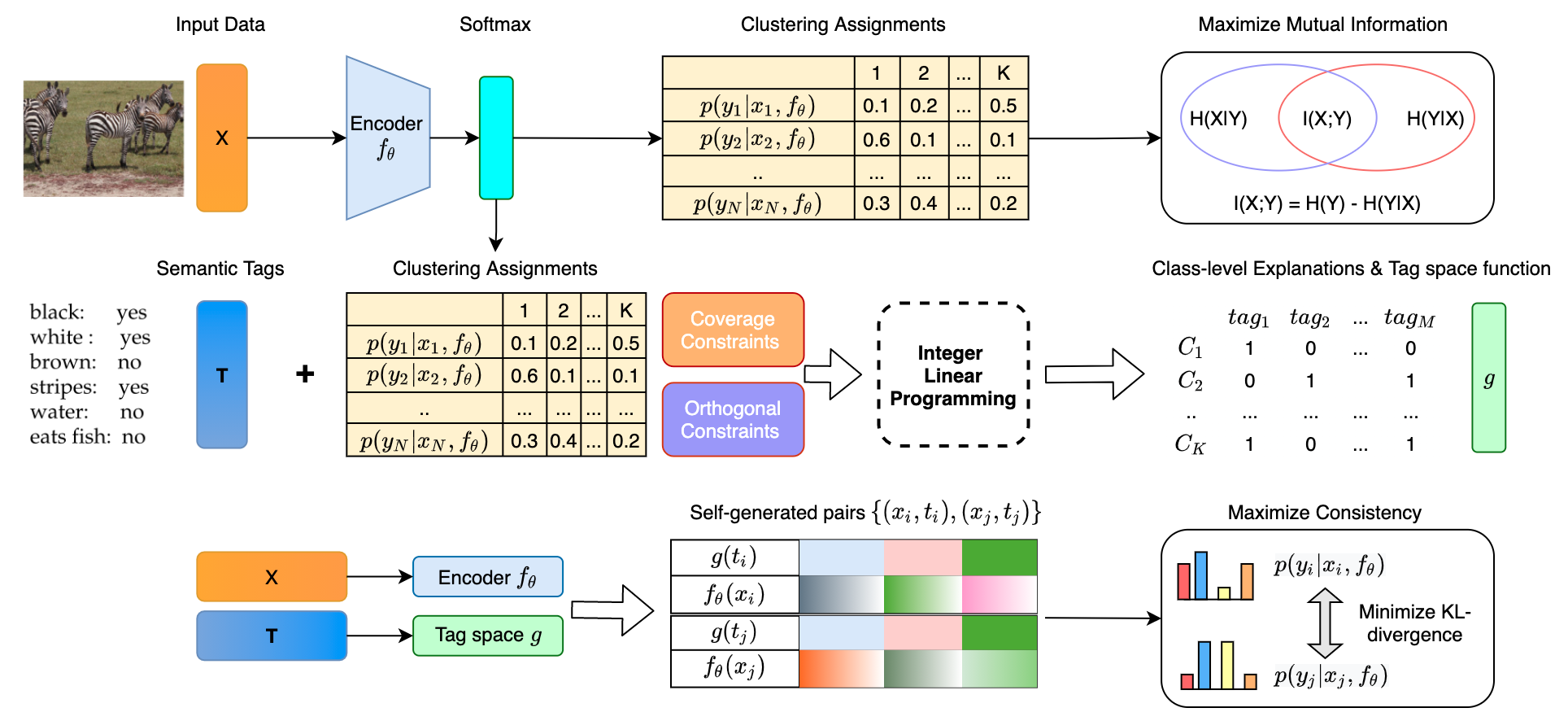}
\caption{The framework of deep descriptive clustering (DDC). DDC consists of one clustering objective, one sub-symbolic explanation objective,  and one self-generated objective to maximize the consistency between clustering and explanation modules. }
 \label{fig:framework}
\end{figure*}
\section{Approach}
\label{sec:approach}
\subsection{Overall Framework}
The framework of our proposed Deep Descriptive Clustering (DDC) is shown in Figure \ref{fig:framework}. 
It can be divided into three learning objectives: i) \emph{clustering objective} which maximizes the mutual information between the empirical distribution on the inputs and the induced label distribution; ii) \emph{class-level explanation objective} which finds the shortest and different explanations for each cluster and creates a tag space mask function $g$ to filter out uninformative tags; iii) \emph{an instance pairwise loss term} with self-generated constraints to maximize the consistency between the clustering feature space and the descriptive tag space induced by mask function $g$.

\subsection{Information Maximization for Clustering}
Given unlabeled dataset of $N$ data points as $X = \{x_1, ..., x_N\}$ where $x_i = (x_{i1}, ..., x_{iD}) \in {\R}^D$ are $D$ dimensional feature vectors, the goal of our proposed model is to predict the clustering assignments $y \in \{1, ..., K\}$ given input $x$, encoding network $f_{\theta}$ and cluster size $K$. Inspired by RIM \cite{krause2010discriminative} which learns a probabilistic clustering model $p(y|x, \W)$ with parameters $\W$ to maximize the mutual information between $x$ and $y$, we represent the estimated mutual information \cite{bridle1992unsupervised} between $x$ and $y$ with network $f_{\theta}$ as the difference between marginal entropy $H(Y)$ and conditional entropy $H(Y{\mid}X)$. Our clustering objective maximizes the mutual information $I(X;Y)$ via minimizing loss $\mathcal{L}_{MI}$:
\begin{equation}
\begin{split}
    \mathcal{L}_{MI} &= -I(X;Y) = H(Y{\mid}X) - H(Y)\\
    &= \frac{1}{N}\sum_{i=1}^{N} h(p(y_i|x_i, f_{\theta})) - h(\frac{1}{N} \sum_{i=1}^{N} p(y_i | x_i, f_{\theta}))
\end{split}
\label{eq:mutual_information}
\end{equation}
where $h$ is the entropy function and $p(y_i | x_i, f_{\theta})$ is calculated through $f_{\theta}$ and $K$-way softmax function. Intuitively minimizing conditional entropy $H(Y{\mid}X)$ will map similar $x$ to have similar clustering predictions $y$ and maximizing the entropy $H(Y)$ will incorporate the notion of class balance to avoid degenerate solution such as all the points map to one class.

\subsection{The Cluster-level Explanation Objective}
In addition to the unlabeled data $X$, to provide high-level explanations we assume a set of partial annotated tags $T = \{t_1, ... t_N\}$ where $t_i = (t_{i1}, ..., t_{iM}) \in {\R}^{M}$ is a binary vector. In real world applications assuming each instance has a complete set of annotated tags is unlikely, thus we assume each instance's tag can be missing with a specified probability $r$. With the predicted clustering assignments $Y$ we can partition both $X$ and $T$ into $K$ clusters. 

We formulate the cluster-level description problem as an Integer Linear Programming (ILP) to learn short and orthogonal descriptors for each cluster. Specifically, we solve for the $K \times M$ binary allocation matrix $W$ where $W_{i,j} = 1$ iff cluster $C_i$ is described by tag $j$. The main objective function is to find the most concise overall cluster description:
\begin{equation}
    \argmin_{W} \sum_{i,j} W_{ij}
    \label{eq:ilp1}
\end{equation}

Our first constraint set includes the explanation length requirement for each cluster explicitly and set coverage constraints implicitly. Given a fixed number of tags as explanations for discovered cluster $C_i$, a high coverage explanation indicates that most instances within cluster $C_i$ contain the selected tags. 
Now we define 
the fraction of cluster $i$ having tag $j$ as $Q_{ij} = \frac{1}{|C_i|} \sum_{t_{k} \in C_i} t_{kj}$. Note we use mean imputation for missing tags. 
Formally we expect at least $\alpha$ tags being selected to explain each cluster:
\begin{equation}
    \sum_{j=1}^{M} W_{ij} Q_{ij} \ge \alpha \quad  \forall i \in \{1, .. K\}
    \label{eq:ilp2}
\end{equation}
Combining Eq (\ref{eq:ilp2}) with our main objective in Eq (\ref{eq:ilp1}), the constraint set in Eq (\ref{eq:ilp2}) will naturally require the ILP solver to select tags that have higher coverage within each cluster.

Our next orthogonal constraint requires that the tags chosen to represent each cluster have minimum overlap which encourages informative explanations. Denote the hyper-parameter $\beta$ as the upper-bound of expected number of overlapping tags per cluster, the orthogonal constraint can be encoded as follow:  
\begin{equation}
    \sum_{i=1}^{K} W_{ij} Q_{ij} \le \beta \quad  \forall j \in \{1, .. M\}
    \label{eq:ilp3}
\end{equation}
Lastly we have the regular constraints to make $W$ a valid binary matrix as $W_{ij} \in \{0, 1\}$. There are $KM$ variables to solve and $K+M$ constraints for set coverage and orthogonal requirements. Our proposed ILP objective can be solved efficiently due to the cluster-level explanation design ($K \ll N $). Empirical results have shown that our ILP module's running time only takes $1\%$ of the whole framework's training time. 

Now we define the tag space mapping function $g$ which is used in our next objective. Let the solution for our proposed cluster-level explanation problem be $W^{*}$. We define function $g$ for all the data as $g(t_i)  = t_i * G$ where $G \in {\R}^{M \times M}$ is a diagonal matrix such that $G_{jj} = 1$ iff tag $j$ is used in explanation allocation matrix $W^{*}$. Note function $g$ can be treated as a mask function to filter out less informative semantic tags solved by our proposed cluster-level explanation objectives.

\subsection{Self-generated Pairwise Loss Term}
Our first proposed objective trains network $f_{\theta}$ for clustering; the second objective solves for explanations and a tag space function $g$. We propose a pairwise loss objective to reconcile inconsistencies between the two by finding instances that share similar informative descriptors but from different clusters, that is $g(t_i) \approx g(t_j)$ but $f_{\theta}(x_i) \neq f_{\theta}(x_j)$.
To achieve this we introduce a pairwise loss objective to bridge the explanation and clustering module. This part is important because our goal is to use semantic tags to generate explanations and reshape the clustering features for better clustering. 
Previous works on constrained clustering \cite{wagstaff2001constrained,basu2008constrained} have shown that adding pairwise guidance such as \emph{together} and \emph{apart} constraints to clustering modules can largely improve clustering performance. However, these algorithms assume the pairwise guidance is given as ground-truth. In our setting we propose to add self-generated pairwise constraints with the assumption that instances which are close in tag space should be close in clustering feature space. Formally for each instance $x_i$ we search for top $l$ instances which minimize the objective $J$ for self-generated \emph{together} constraints: 
\begin{equation}
\begin{split}
    J_i = & \min_{j \in \{1, ... , N\}} \gamma*|g(t_i) - g(t_j)| - |f_{\theta}(x_i) - f_{\theta}(x_j)|
\end{split}
\label{eq:bi}
\end{equation}
 where $\gamma$ is the penalizing weight for tag space's difference. Minimizing $J$ requires accessing the whole training set which is inefficient for mini-batch training. Instead we replace $N$ with batch size $N_B$ and solve an approximated version of Eq (\ref{eq:bi}) in each mini-batch. We generate $l$ pairs of together constraints for each instance $x_i$ and then directly minimize the KL divergence between the clustering predictions $y_i$ and $y_j$:
\begin{equation}
\begin{split}
        \mathcal{L}_{P} =  & \frac{1}{Nl} \sum_{i=1}^{N} \sum_{j=1}^{l} KL(p(y_i | x_i, f_{\theta}), p(y_j | x_j, f_{\theta})) \\
\end{split}
\label{eq:pairwise}
\end{equation}
Eq (\ref{eq:pairwise}) minimizes the inconsistency between the clustering feature space and the semantic tag space and reshapes the clustering feature space for better clustering and explanation.

\renewcommand{\algorithmicrequire}{\textbf{Input:}}
\renewcommand{\algorithmicensure}{\textbf{Output:}}
\begin{algorithm}[th]
 \caption{Algorithm for Deep Descriptive Clustering}
 \label{alg:1}
\begin{algorithmic}[1]
\REQUIRE Data $X = \{x_1, ..., x_N\}$, tags $T = \{t_1, ..., t_N\}$, number of clusters $K$, hyper-parameters $\alpha,\beta,\gamma,\lambda$.
\ENSURE Clustering partitions $\{ C_1, ... C_K\}$, well-trained $f_{\theta}$ and $g$, explanation allocation matrix $W^{*}$.
 \STATE Initialize network $f_{\theta}$ and tag space function $g$.
 \STATE Pre-train $f_{\theta}$ via back-propagating overall loss in Eq (\ref{eq:overall_loss}).
 \REPEAT
 \STATE Construct cluster-level explanation problem as ILP defined in Eq (\ref{eq:ilp1},\ref{eq:ilp2},\ref{eq:ilp3}). Initialize $\beta = 0$, $W^{*} = \emptyset$.
 \WHILE{ILP solution $W^{*}$ is not feasible}
 \STATE Increase hyper-parameter $\beta$ by the fixed step size $1$. 
 \STATE Solve the ILP for $W^{*}$ and tag space function $g$.
 \ENDWHILE
 \FOR{each mini-batch}
 \STATE Generate pairwise constraints based on the objective $J$ in Eq (\ref{eq:bi}) within each batch.
 \STATE Calculate the pairwise loss $\mathcal{L}_P$ via Eq (\ref{eq:pairwise}) and the clustering loss $\mathcal{L}_{MI}$ via Eq (\ref{eq:mutual_information}).
 \STATE Update network parameters $f_{\theta}$ by minimizing overall loss $\mathcal{L}$ in Eq (\ref{eq:overall_loss}).
 \ENDFOR
\UNTIL{Network $f_{\theta}$ and explanation results converge}
\end{algorithmic}
\end{algorithm}

\subsection{Overall Training Algorithm}
Algorithm \ref{alg:1} presents our training algorithm for the deep descriptive clustering. Firstly we initialize the clustering network $f_{\theta}$ with random weights and initialize the weight matrix $G$ of function $g$ as identity matrix. Then we minimize the overall loss $\mathcal{L}$ by combining the clustering objective $\mathcal{L}_{MI}$ and pairwise loss term $\mathcal{L}_{P}$ with weight $\lambda$:
\begin{equation}
\begin{split}
    \mathcal{L} & = \frac{\lambda}{Nl}  \sum_{i=1}^{N} \sum_{j=1}^{l} KL (p(y_i | x_i, f_{\theta}), p(y_j | x_j, f_{\theta})) +\\
    &\frac{1}{N}\sum_{i=1}^{N} h(p(y_i|x_i, f_{\theta})) -  
    h(\frac{1}{N} \sum_{i=1}^{N} p(y_i | x_i, f_{\theta}))
    \label{eq:overall_loss}
\end{split}
\end{equation}
Given the clustering predictions we construct the cluster-level explanation problem with binary variable $W$ and calculate $Q$ values for all the discovered clusters $\{C_1, ... C_K\}$. Note given the expected number of tags used for each cluster as $\alpha$, we run our ILP solver iteratively with linear search for the smallest feasible hyper-parameter $\beta$ to ensure tightest orthogonal constraints. Once the binary explanation allocation matrix $W^{*}$ is solved, we update the tag space function $g$ and regenerate the pairwise constraints via objective $J$ to maximize the consistency between clustering features and tag space. The whole model is trained repetitively until convergence.  
 \begin{table*}[th]
  \centering
  \begin{tabular}{lp{8.5cm}p{5cm}cc}
  \hline
    C & Composition by animals & Description by tags &TC $\uparrow$ &ITF $\uparrow$ \\
    \hline
    C1 & 1 grizzly bear, 2 dalmatian, 1 horse, 2 blue whale & big, fast, strong, muscle, new world, smart & 0.94 & 1.34\\
    \hline
    C2 & 5 antelope, 2 grizzly bear, 2 beaver, 5 dalmatian, 5 Persian cat, 5 horse, 6 German shepherd, 3 Siamese
cat & furry, chew teeth, fast, quadrupedal, new world, ground & 0.98 & 0.94\\
    \hline
    C3 & 69 beaver, 64 dalmatian, 42 Persian cat, 29 blue whale, 42 Siamese cat & tail, fast, new world, timid, smart, solitary & 0.98 & 1.17\\
    \hline
    C4 & 100 killer whale, 69 blue whale, 1 Siamese cat & tail, fast, fish, smart & 1.00 & 1.10\\
    \hline
    C5 & 95 antelope, 97 grizzly bear, 29 beaver, 29 dalmatian, 53 Persian cat, 94 horse, 94 German shepherd, 54
Siamese cat & furry, chew teeth, fast, quadrupedal, new world, ground & 1.00 & 0.94\\
    \hline
    \end{tabular}
    \caption{Results generated by descriptive clustering \protect \cite{dao2018descriptive}, we present the first Pareto point of their result  such that the diameter of all the clusters are minimized. $\uparrow$ means the larger value the better. }
    \label{tab:dc}
\end{table*}

 \begin{table*}[th]
  \centering
  \begin{tabular}{lp{4.5cm}p{9cm}cc}
  \hline
    C & Composition by animals & Description by tags  &TC $\uparrow$ &ITF $\uparrow$\\
    \hline
    C1 & 100 grizzly bear, 100 beaver& tough skin, bulbous, paws, quadrupedal, nocturnal, hibernate, smart, solitary & 1.00 & 2.32\\
    \hline
    C2 & 100 Siamese cat, 100 Persian cat & white, gray, pads, chew teeth, claws, weak, inactive, old world & 1.00 & 2.32\\
    \hline
    C3 & 100 antelope, 100 dalmatian & furry, big, long leg, active, new world, ground, timid, group & 1.00 & 2.32\\
    \hline
    C4 & 100 killer whale, 100 blue whale& spots, hairless, flippers, strain teeth, fish, plankton, arctic, ocean & 1.00 & 2.32\\
    \hline
    C5 & 100 horse, 100 German shepherd& black, brown, patches, smelly, walks, strong, agility, domestic & 1.00 & 2.32\\
    \hline
\end{tabular}
  \caption{ Results generated by our proposed DDC. $\uparrow$ means the larger value the better.}
    \label{tab:ddc}
\end{table*}

\section{Experiments}
\label{sec:experiments}
In this section, we conduct experiments to evaluate our approach empirically. Based on our experiments, we aim to answer the following questions:
\begin{itemize}
 \item Can our proposed approach generate better explanations compared to existing methods? (see Sec \ref{sec:exp1}) Can it generate more complex explanations such as ontologies (see Sec \ref{sec:exp2})?
\item How does our proposed approach perform in terms of clustering quality? (see Sec \ref{sec:cluster})
\item How does simultaneously clustering and explaining improve our model's performance? (see Sec \ref{sec:ablation})
\end{itemize}

\subsection{Experimental Setup}
\paragraph{Data.} We evaluate the performance of our proposed model on two visual data sets with annotated semantic attributes. We first use Attribute Pascal and Yahoo (aPY) \cite{farhadi2009describing}, a small-scale coarse-grained dataset with $64$ semantic attributes and $5274$ instances. We have selected $15$ classes for our clustering task. Further, we have studied Animals with Attributes (AwA) \cite{lampert2013attribute}, which is a medium-scale dataset in terms of the number of images. For AwA we use $85$ semantic attributes and $19832$ instances, we have set $40$ classes for clustering, the total number of animals. 

\paragraph{Baselines and Evaluation Metrics.} In the experiments, deep descriptive clustering is compared with descriptive clustering \cite{dao2018descriptive} in terms of the explanation quality. To evaluate the generated explanations quantitatively and qualitatively, we list all the composition and selected tags for each discovered cluster and report the \emph{Tag Coverage} (TC) and \emph{Inverse Tag Frequency} (ITF). For cluster $C_i$, let the descriptive tag set be $D_i$, the TC and ITF for $C_i$ are calculated as:
\begin{equation}
    TC(C_i) = \frac{1}{|D_i|} \sum_{d \in D_i} \frac{|\{ (x,t) \in C_i : d \in t\}|}{|C_i|}
\end{equation}
\begin{equation}
    ITF(C_i) = \frac{1}{|D_i|} \sum_{d \in D_i} \log \frac{K}{\sum_{j=1}^K |d \in D_j|}
\end{equation}
The \emph{Tag Coverage} for $C_i$ ranges from $[0,1]$ and the max value is achieved when each descriptive tag exists in all the instances within $C_i$. The \emph{Inverse Tag Frequency} for $C_i$ ranges from $[0, \log K]$ and the max value is achieved when each descriptive tag is only used once. For both TC and ITF the \emph{larger} the better. We have also generated a \emph{graphical ontology} as high-level explanation on the clustering results when the number of clusters is large and a long explanation list is problematic. We evaluate its quality by comparing it to human knowledge. 

Further, we evaluate the clustering performance with a range of tag annotated ratios $r$ as [$10 \%, 30 \%, 50 \%$] and compare DCC's results against vanilla k-means clustering and competitive incomplete multi-view clustering approaches such as MIC, IMG, and DAIMC \cite{shao2015multiple,zhao2016incomplete,hu2018doubly}.
For the clustering evaluation metric, we choose to use both Normalized Mutual Information (NMI) \cite{strehl2000impact} and Clustering Accuracy (ACC) \cite{xu2003document} for comprehensive evaluation.

\paragraph{Implementations.} For a fair comparison with all the baseline approaches, we use pre-trained ResNet-101 \cite{he2016deep} features for all the clustering tasks and the encoder networks of deep descriptive clustering model are stacked by three fully connected layers with size
of $[1200, 1200, K]$ where $K$ is the desired number of clusters. We set the expected number of tags for each cluster as $8$ and hyper-parameters $l, \lambda, \gamma$ as $1, 1, 100$ respectively. The tag annotated ratio $r$ is set as $0.5$ by default to simulate a challenging setting. The activation function is ReLU, and the optimizer is Adam \cite{kingma2015adam} with default parameters. 

\subsection{Comparison with Descriptive Clustering}
\label{sec:exp1}
We duplicate the experimental setting in \cite{dao2018descriptive} by down-sampling $10$ classes from AwA and cluster the data into $5$ clusters for a fair comparison. We list the explanation results in Table \ref{tab:dc} and \ref{tab:ddc}. Our model's \emph{Tag Coverage} values for all the clusters are $1$; this result shows that our model successfully maximizes the consistency between the discovered tag space and clustering feature space so that similar instances with similar tags are grouped. Moreover, the \emph{Inverse Tag Frequency} values of our model are much higher than the competing method. This result indicates that our model selects informative tags for each cluster that differentiate from other discovered clusters. We also observe that our proposed model generates high-quality clusters where similar animals are correctly grouped together. Finally, we have found one attribute's annotation error in the AwA data when examining our explanations for $C_1$; the beavers are annotated with attribute \emph{hibernate} but the truth is the opposite. This finding suggests that the labeled attributes are noisy in the AwA data set.

 \begin{table*}[th]
  \centering
  \begin{tabular}{|c|c|ccc|ccc|}
  \hline
    \multicolumn{2}{|c|}{}&\multicolumn{3}{|c|}{NMI}&\multicolumn{3}{|c|}{ACC} \\
    \hline
    Datasets &Methods $ r\%$ &10 & 30 & 50  &10 & 30 & 50  \\
    \hline
    \multirow{5}{*}{\rotatebox[origin=c]{0}{AwA} }&K-Means& $\text{71.67} \pm \text{0.63}$ & $\text{73.72} \pm \text{0.66}$  &$\text{74.23} \pm \text{0.69}$ &$\text{66.21} \pm \text{0.57}$ & $\text{67.98} \pm \text{0.60}$   &$\text{68.24} \pm \text{0.54}$\\
    &IMG    & $\text{71.86} \pm \text{2.41}$ &$\text{74.43} \pm \text{2.69}$  &$\text{82.16} \pm \text{3.01}$  &$\text{66.19} \pm \text{2.05}$  &$\text{69.17} \pm \text{2.25}$  &$\text{76.24} \pm \text{2.78}$\\
    &MIC    & $\text{72.40} \pm \text{1.68}$ &$\text{76.85} \pm \text{1.71}$  &$\text{83.43} \pm \text{1.89}$&$\text{67.26} \pm \text{1.45}$  &$\text{70.52} \pm \text{1.58}$  &$\text{77.68} \pm \text{1.84}$\\
    &DAIMC  & $\text{72.88} \pm \text{2.38}$ &$\text{79.02} \pm \text{2.46}$  &$\text{87.10} \pm \text{2.74}$  &$\text{67.87} \pm \text{1.97}$  &$\text{73.14} \pm \text{2.13}$   &$\text{82.34} \pm \text{2.39}$\\
    &Ours DCC& $\textbf{75.62} \pm \textbf{1.17}$  &$\textbf{83.93} \pm \textbf{1.35}$  &$\textbf{89.57} \pm \textbf{1.37}$&$\textbf{71.19} \pm \textbf{0.93}$  &$\textbf{78.74} \pm \textbf{1.12}$  &$\textbf{84.48} \pm \textbf{1.20}$\\
    \hline
    \multirow{5}{*}{\rotatebox[origin=c]{0}{aPY}}&K-Means&\text{63.08} $\pm$ \text{0.45} & \text{63.89} $\pm$ \text{0.42}  &\text{64.38} $\pm$ \text{0.48} &\text{57.11} $\pm$ \text{0.39} &\text{58.13} $\pm$ \text{0.36}   &\text{58.98} $\pm$ \text{0.37}\\
    &IMG    &\text{64.75} $\pm$ \text{2.05} &\text{70.19} $\pm$ \text{2.19}  &\text{77.50} $\pm$ \text{2.37} &\text{60.18} $\pm$ \text{1.78}  &\text{65.72} $\pm$ \text{1.90}  &\text{71.21} $\pm$ \text{1.96}\\
    &MIC    & \text{65.36} $\pm$ \text{1.49} &\text{73.89} $\pm$ \text{1.61}  &\text{80.38} $\pm$ \text{1.83}  &\text{62.36} $\pm$ \text{1.28}  &\text{66.98} $\pm$ \text{1.40}  &\text{72.42} $\pm$ \text{1.53}\\
    &DAIMC  & \text{69.29} $\pm$ \text{1.82} &\text{80.70} $\pm$ \text{1.91}  &\text{84.24} $\pm$ \text{1.97}&\text{68.21} $\pm$ \text{1.54}  &\text{73.63} $\pm$ \text{1.63} &\text{76.11} $\pm$ \text{1.68}\\
    &Ours DCC& \textbf{70.54} $\pm$ \textbf{0.98} &\textbf{82.41} $\pm$ \textbf{1.15}  &\textbf{86.30} $\pm$ \textbf{1.22}&\textbf{69.30} $\pm$ \textbf{0.86}& \textbf{76.34} $\pm$ \textbf{0.95}  &\textbf{79.87} $\pm$ \textbf{1.02}\\
    \hline
\end{tabular}
  \caption{Comparison of clustering performance averaged over $10$ trials (mean $\pm$ var) on AwA and aPY under different tag annotated ratio $r \% \in \{10, 30, 50\}$. Bold results are the best mean results among all the algorithms.}
  \label{tab:clustering}
\end{table*}

\subsection{Novel Explanation as Ontology Extraction}
\label{sec:exp2}
\begin{figure}[t]
\vskip -0.2in
\centering
 \includegraphics[width=0.45\textwidth]{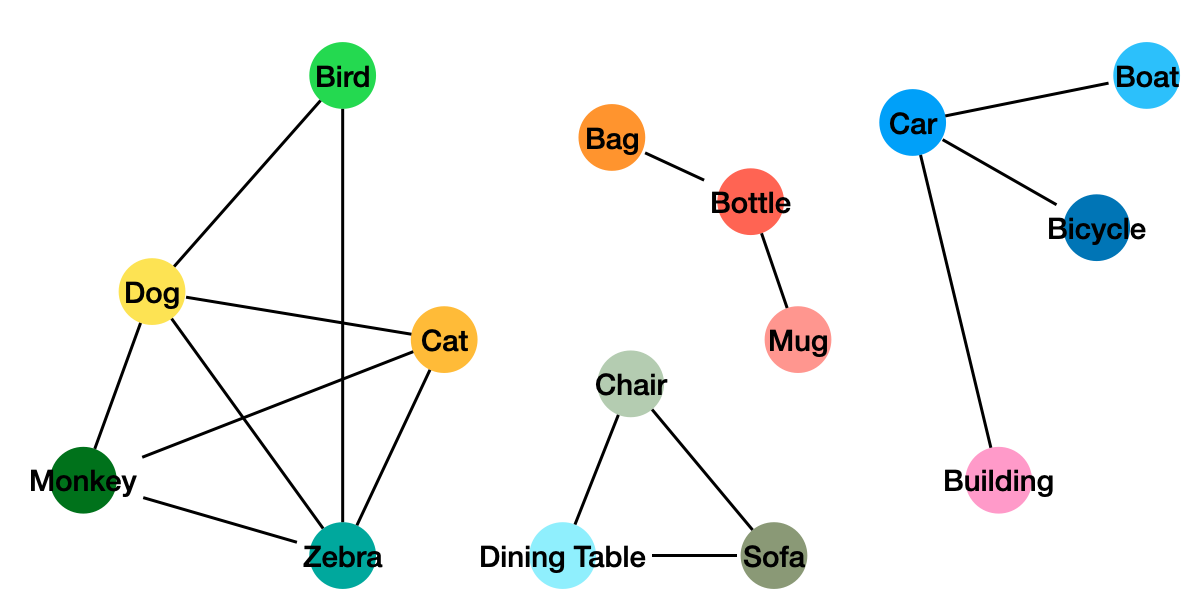}
\caption{The graphical ontology generated for aPY data set.}
 \label{fig:aPY}
 \vskip -0.05in
\end{figure}

Interpreting the descriptions for each cluster can be problematic when the number of clusters is large and the description list is long. We propose to generate a graphical ontology that not only describes each cluster but shows the relationships between them to better inform people. We have visualized the ontology graph for aPY in Figure \ref{fig:aPY}. The nodes represent discovered clusters and the name of the nodes corresponds to the majority class within each cluster. When two clusters share at least $q$ tags ($q=3$ in our experiments) we add an edge between these two clusters. This shows a higher level of structure as we can see the ontology plot in Figure \ref{fig:aPY} reflects four distinct communities which are animals, transportation tools, furniture, and small objects. 
Those ontologies are generally in line with human knowledge and provide a high-level abstraction explanation of our deep descriptive clustering model. 

\subsection{Evaluating Clustering Performance}
\label{sec:cluster}
Here we report if the descriptive clustering problem can increase clustering quality. Since these methods are not deep learning based, to make a fair comparison we use the same pre-trained ResNet-101 features. We report the clustering results of our model under a range of annotated ratios in Table \ref{tab:clustering}. We have several observations to highlight: firstly our model consistently outperforms the k-means and multi-view clustering baselines with different tag annotated ratios; secondly with more annotated tags, both multi-view clustering baselines and our model improves largely comparing to the k-means clustering which naively concatenates the images features with semantic attributes. We attribute the good clustering performance for both deep representation learning and our novel way of leveraging semantic tag information for better clustering. 

 \begin{figure}[t]
 \vskip -0.15in
 \centering
 \subfloat[Parameter analysis on $\alpha$]{
 \includegraphics[width=0.235\textwidth]{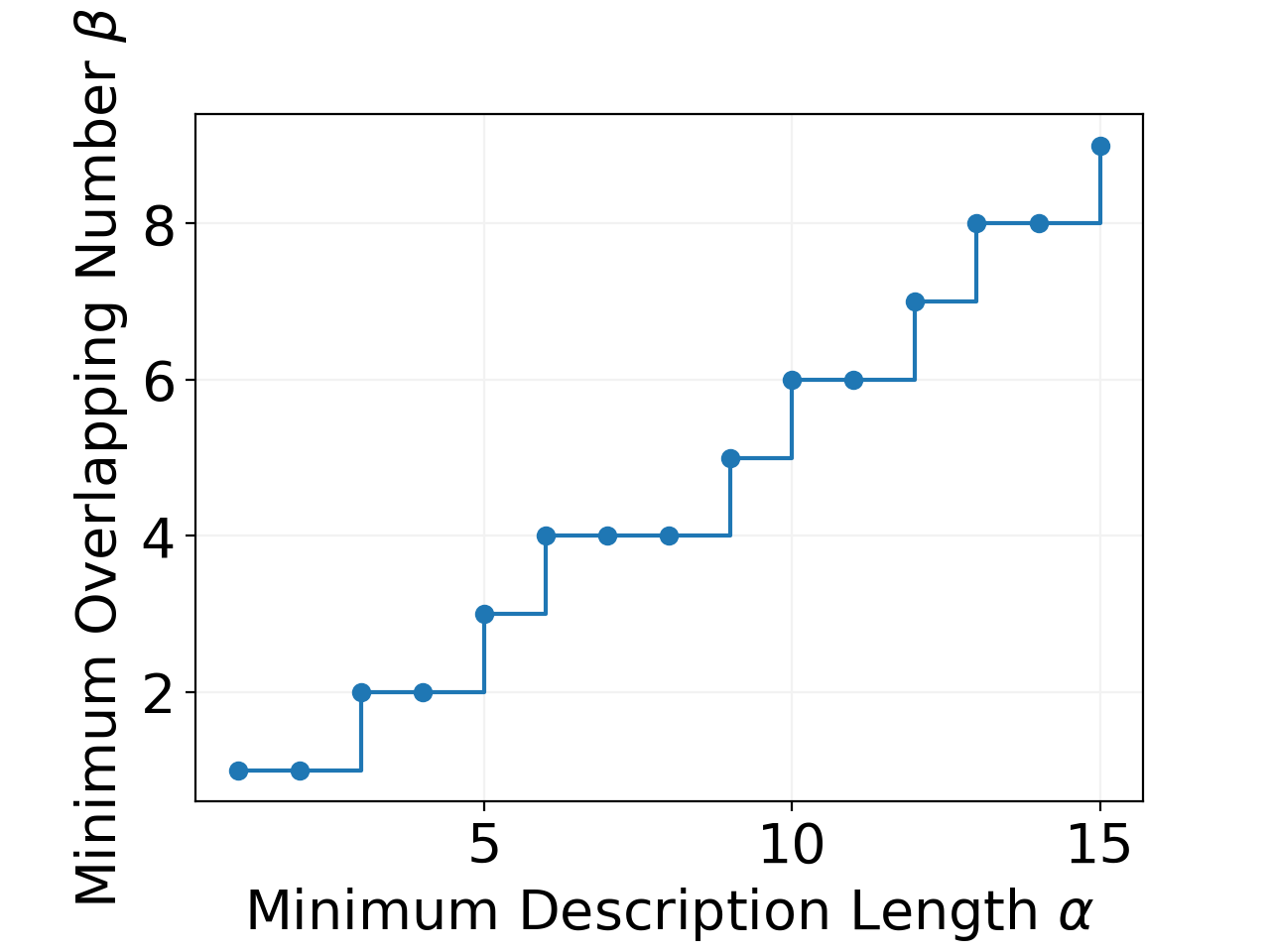}}
 \hfill
 \subfloat[Ablation study on function $g$ ]{
 \includegraphics[width=0.235\textwidth]{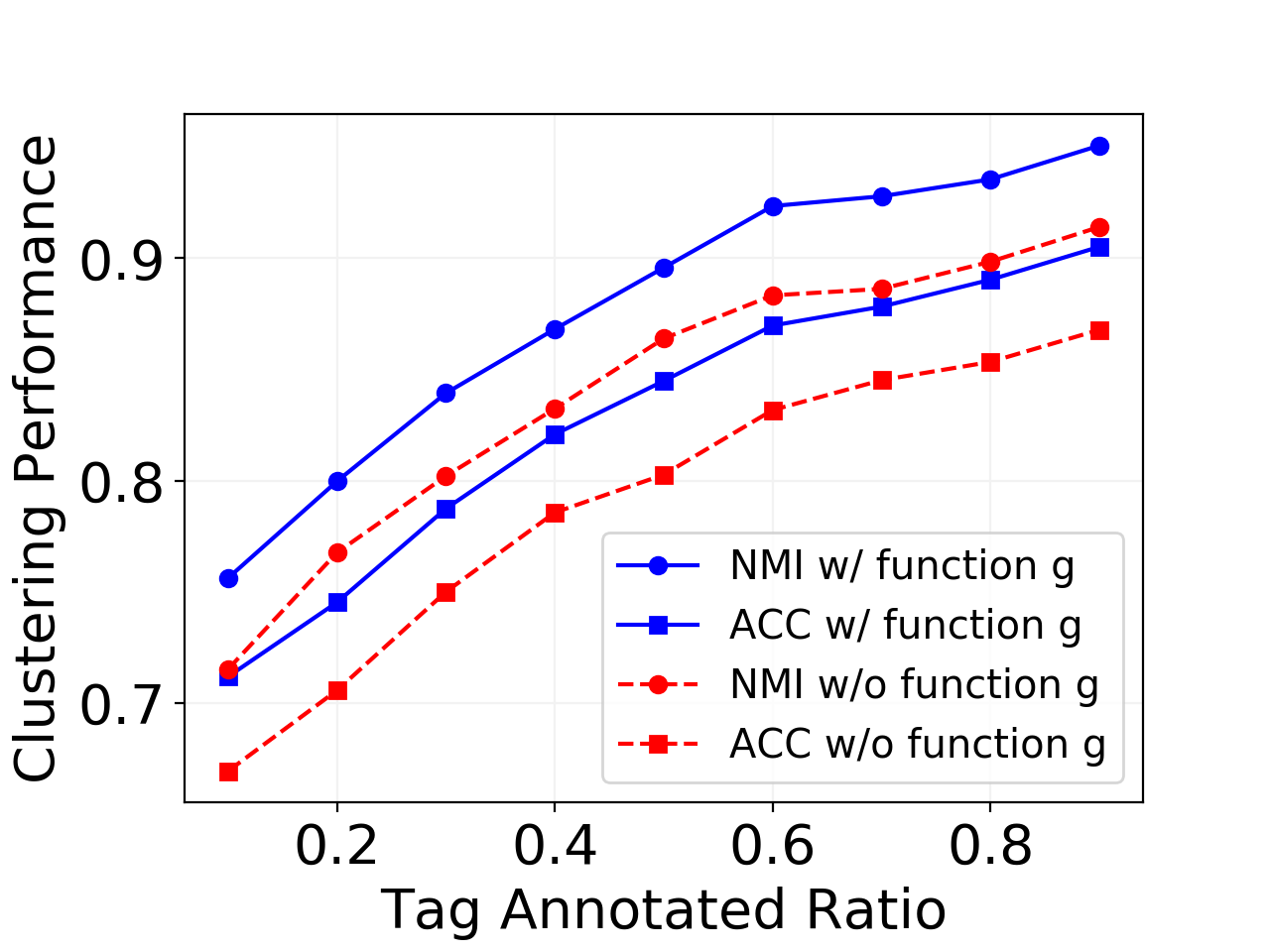}}
\caption{Plots for parameter analysis and ablation study}
\label{fig:analysis}
 \vskip -0.05in
\end{figure}

\subsection{Parameter Analysis and Ablation Test}
\label{sec:ablation}
Given the hyper-parameter $\alpha$ which denotes the minimum expected number of tags for description, we plot the automatically searched parameter $\beta$ for AwA in Figure \ref{fig:analysis} (a). This result shows our automatic searching procedure's success and suggests that a relatively small $\alpha$ leads to more concise and orthogonal explanations. Meanwhile, we conduct ablation experiments to analyze the impact of mask function $g$ solved via our explanation module. In Figure \ref{fig:analysis} (b), the blue lines indicate clustering results with function $g$. In red lines we replace function $g$ with an identity function to conduct the ablation study. Comparing red and blue lines we can see that mask function $g$ can remove the noisy information within semantic tag space and consistently improve the clustering performance.

\section{Conclusion and Future Work}
\label{sec:conclusion}
This paper proposes deep descriptive clustering, which can learn to cluster and generate cluster-level explanations simultaneously. We develop a novel deep learning framework that supports learning from the sub-symbolic level (which clustering is performed on) and symbolic level (which explanations are created from). Empirical results on real-world data demonstrate the high quality of our generated explanations and good clustering performance. Our future work will focus on building an explainable clustering model with noisy semantic features and exploring other novel forms of explanations beyond ontologies on different types of data.

\small
\bibliographystyle{named}
\bibliography{ijcai21}

\newpage

\end{document}